\documentclass[conference]{IEEEtran}
\IEEEoverridecommandlockouts
\usepackage{cite}
\usepackage{amsmath,amssymb,amsfonts}
\usepackage{algorithmic}
\usepackage{graphicx}
\usepackage{textcomp}
\usepackage{xcolor}
\usepackage[a4paper, total={184mm,239mm}]{geometry}
\def\BibTeX{{\rm B\kern-.05em{\sc i\kern-.025em b}\kern-.08em
    T\kern-.1667em\lower.7ex\hbox{E}\kern-.125emX}}

\usepackage{mathtools}

\usepackage{cite}
\usepackage{amsmath,amssymb,amsfonts}
\usepackage{algorithmic}
\usepackage{graphicx}
\usepackage{textcomp}
\usepackage{xcolor}
\usepackage{threeparttable}
\def\BibTeX{{\rm B\kern-.05em{\sc i\kern-.025em b}\kern-.08em
    T\kern-.1667em\lower.7ex\hbox{E}\kern-.125emX}}
\usepackage{multirow}
\usepackage[caption=false]{subfig}
\usepackage{tabu}
\usepackage{color, colortbl}
\newcommand{\thickhline}{%
    \noalign {\ifnum 0=`}\fi \hrule height 1pt
    \futurelet \reserved@a \@xhline
}
\usepackage[linesnumbered,ruled]{algorithm2e}
\usepackage{xcolor}

\SetCommentSty{mycommfont}    
\usepackage{pifont}
\usepackage{cite}
\usepackage{amsmath}
\newtheorem{Tlemma}{Lemma}

\newtheorem{Tdef}{Definition}
\newenvironment{Definition}[1]
    {\begin{Tdef}\noindent\textsc{(#1)}\itshape}
    {\end{Tdef}}
\usepackage{amsmath}

\newcommand{\ineq}[1]{\footnotesize$#1$\normalsize}{}
{}

\newcommand{\mubrain}{$\mu$\text{{Brain}}}{}
\newcommand{\sm}{\text{{SpiNeMap}}}{}
\newcommand{\tech}{\text{{SentryOS}}}{}
\newcommand{\techcompile}{\text{{SentryC}}}{}
\newcommand{\techrt}{\text{{SentryRT}}}{}
\DeclareMathAlphabet{\mymathbb}{U}{BOONDOX-ds}{m}{n}

\begin{document}
\bstctlcite{IEEEexample:BSTcontrol}
\title{\fontsize{23pt}{30pt}\selectfont Design of Many-Core Big Little \mubrain{s} for Energy-Efficient Embedded Neuromorphic Computing}

\author{\IEEEauthorblockN{M. L. Varshika${}^\dagger$, Adarsha Balaji${}^\dagger$, Federico Corradi${}^\ddagger$, Anup Das${}^\dagger$, Jan Stuijt${}^\ddagger$ and Francky Catthoor${}^{\dagger\dagger}$}
\IEEEauthorblockA{\textit{${}^{\dagger}$Electrical and Computer Engineering, Drexel University, USA} \\
\textit{${}^{\ddagger}$Stichting IMEC Nederland, Netherlands}\\
\textit{${}^{\dagger\dagger}$IMEC, Belgium}\\
Corresponding Email: lm3486@drexel.edu}
}


\maketitle

\begin{abstract}
As spiking-based deep learning inference applications are increasing in embedded systems, these systems tend to integrate neuromorphic accelerators such as \mubrain{} to improve energy efficiency.
We propose a \mubrain{}-based scalable many-core neuromorphic hardware design to accelerate the computations of spiking deep convolutional neural networks (SDCNNs).
To increase energy efficiency, cores are designed to be heterogeneous in terms of their neuron and synapse capacity 
(big cores have higher capacity than the little ones), and they are interconnected using a parallel segmented bus interconnect, which leads to lower latency and energy compared to a traditional mesh-based Network-on-Chip (NoC). We propose a system software framework called \tech{} to map SDCNN inference applications to the proposed 
design. \tech{} consists of a compiler and a run-time manager. The compiler compiles an SDCNN application into sub-networks by exploiting the internal architecture of big and little \mubrain{} cores. The run-time manager schedules these sub-networks onto cores and pipeline their execution to improve throughput. 
We evaluate the proposed big little 
many-core neuromorphic design and the system software framework with five commonly-used SDCNN inference applications and show that the proposed solution reduces energy (between 37\% and 98\%), reduces latency (between 9\% and 25\%), and increases application throughput (between 20\% and 36\%). We also show that 
\tech{} 
can be easily extended for other spiking neuromorphic accelerators.
\end{abstract}


\section{Introduction}\label{sec:introduction}
Spiking deep convolutional neural network (SDCNN)-based inference applications are increasing in embedded systems~\cite{cao2015spiking}. To improve energy efficiency, such systems tend to integrate neuromorphic accelerators such as \mubrain{}~\cite{mubrain}, DYNAPs~\cite{dynapse}, and Loihi~\cite{loihi}. 
We take the example of \mubrain{}, which is a neural architecture with three layers:
16 neurons in the first layer, 64 neurons in the second, and 256 neurons in the third (see Fig.~\ref{fig:mubrain_architecture}). \mubrain{} is an asynchronous (clock-less) digital design with fully programmable connections between the three layers. \mubrain{} is shown to consume only 308nJ energy for 
digit recognition~\cite{mubrain}. 

\mubrain{} implements 336 neurons and 38K synaptic connections. 
One way to use \mubrain{} for larger SDCNN applications 
is to scale-up the design. However, this leads to a substantial increase in static power and area (see Section~\ref{sec:background}). We propose a scalable solution by interconnecting many small 
\mubrain{} cores. However, instead of using the same neuron and synapse capacity for all cores as in all previous designs, we show that a heterogeneous architecture with different core capacities can improve energy efficiency of the proposed many-core design.

Conventionally, mesh-based Network-on-Chip (NoC) is used to 
interconnect cores in recent neuromorphic designs~\cite{yang2020recent}.
However, NoC interconnect has relatively long time-multiplexed connections that need to be near-continuously powered up and down, reaching from the ports of data producers/consumers (inside a core or between 
different cores) up to the ports of communication switches~\cite{wasif2020energy,neunoc,yoon2017system}. 
Recently, segmented bus (SB)-based interconnect is proposed as an alternative for neuromorphic hardware~\cite{balaji2019exploration}. Here, a bus lane is partitioned into segments, where interconnections between segments are bridged and controlled by switches~\cite{chen1999segmented}. 
We propose a dynamic segmented bus architecture with multiple segmented bus lanes to interconnect big little \mubrain{} cores in the proposed design (see Figure~\ref{fig:many_core_mubrain}). 
An optimized controller is designed to perform mapping of communication primitives to segments by profiling the communication pattern between different \mubrain{} cores for a given SDCNN application. Based on this profiling and mapping, switches in the interconnect are programmed once at design-time before admitting an application to the hardware. By avoiding run-time routing decisions, the proposed design significantly gains on energy and latency.


We propose \tech{}, a system software for mapping SDCNN applications to the proposed segmented bus-based many-core big little \mubrain{} design. \tech{} consists of a compiler (\techcompile{}) and a run-time manager (\techrt{}). \techcompile{} compiles an SDCNN inference application into sub-networks by exploiting the internal architecture of big and little \mubrain{} cores. \techrt{} uses a dataflow analysis technique to schedule sub-networks to \mubrain{} cores by improving opportunities for pipelining and exploiting data-level parallelism.

We show in Section~\ref{sec:sentry_ss} that \tech{} can be easily extended
to other many-core spiking neuromorphic designs such as DYNAPs~\cite{dynapse} (which is similar to \mubrain{} with synaptic memory integrated closer to neuron circuitry in each core, but with more neurons and synapses per core) and Loihi~\cite{loihi} (where synaptic memory is off-chip to neuron circuity).

Following are our \textbf{contributions} to the neuromorphic field.
\begin{itemize}
    \item A many-core neuromorphic platform template design based on the digital asynchronous (clock-less) \mubrain{} architecture. Cores in the proposed design are heterogeneous in terms of their neuron and synapse capacity. The main objective here is to reduce energy (Section~\ref{sec:mubrain_many_core}).
    \item A parallel segmented bus-based interconnect for data communication between \mubrain{} cores in the proposed many-core design. A controller to map inter-core communication to segments for parallel execution. The main objective here is to minimize energy and latency (Section~\ref{sec:mubrain_many_core}).
    \item A system software framework (\tech{}) to map SDCNN applications to the proposed design. \tech{} consists of a compiler, which compiles an application into sub-networks and a run-time manager, which schedules sub-networks to \mubrain{} cores of the many-core hardware. The main objective here is to improve throughput (Section~\ref{sec:sentry_ss}).
\end{itemize}

We evaluate the proposed design with five commonly-used SDCNN applications and show improvement in energy (average 67\%), latency (average 18\%), and throughput (average 25\%). 

To the best of our knowledge, this is the first work that proposes a many-core neuromorphic design with heterogeneous core capacity and using a segmented bus interconnect.

\section{Background}\label{sec:background}
\subsection{\mubrain{}: A Digital Inference Hardware}\label{sec:mubrain}
\begin{figure}[h!]%
    \vspace{-10pt}
    \centering
    \subfloat[3-layered \mubrain{} architecture.\label{fig:mubrain}]{{\includegraphics[width=5.0cm]{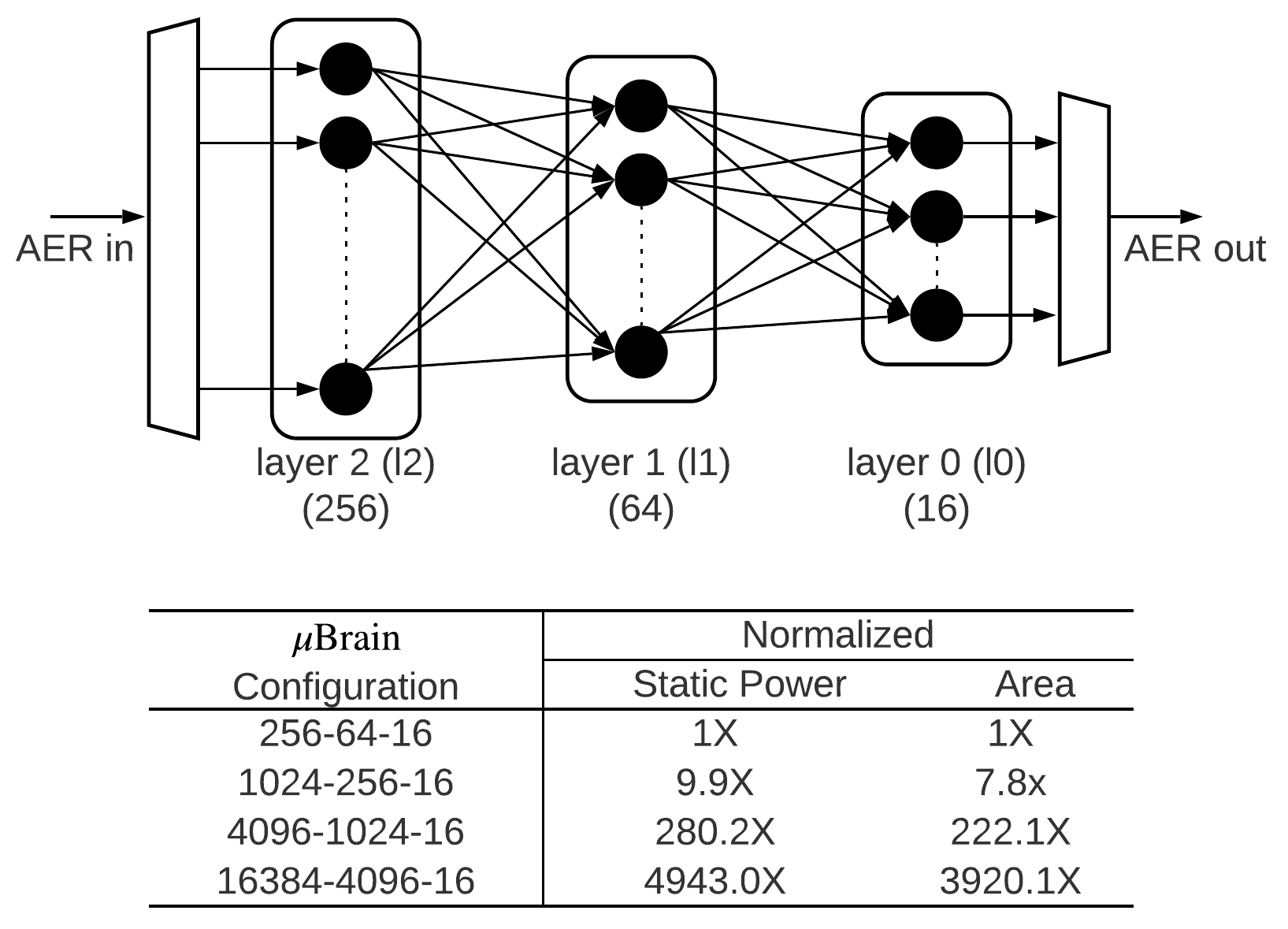} }}%
    \subfloat[State diagram of an IF neuron.\label{fig:state_diagram}]{{\includegraphics[width=4.0cm]{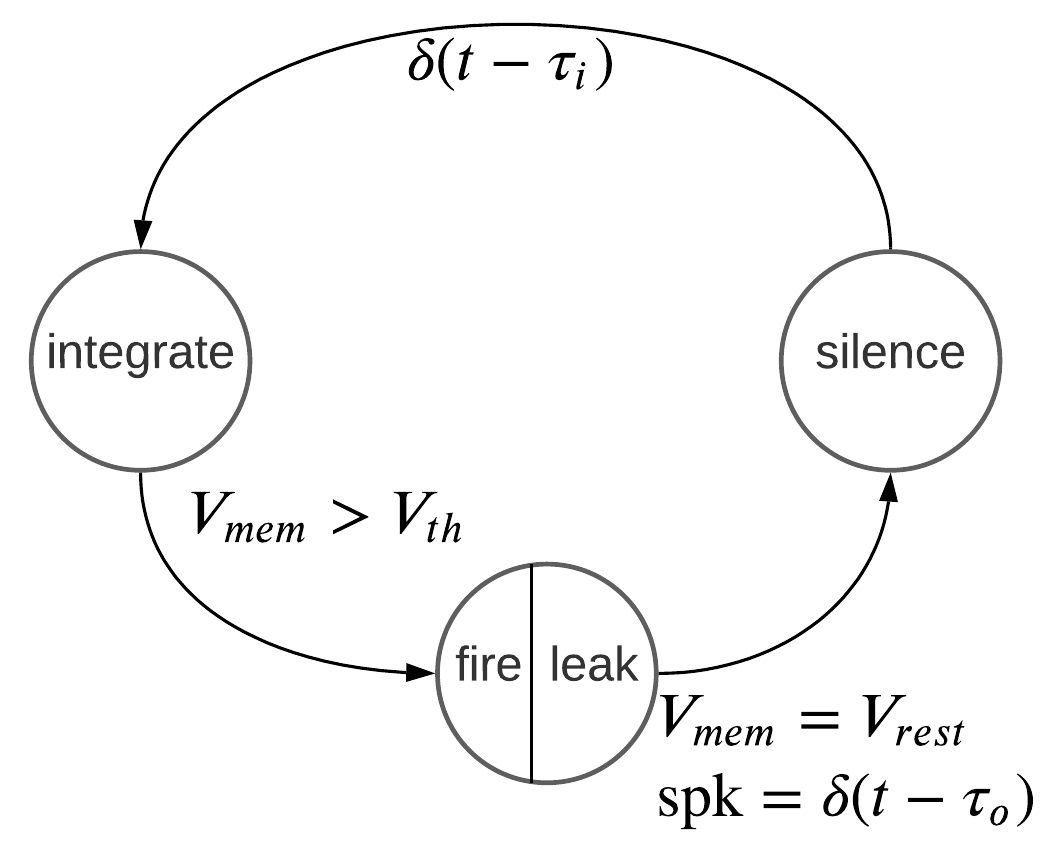} }}%
    \vspace{-5pt}
    \caption{\mubrain{} architecture and neuron state diagram~\cite{mubrain}.}%
    \label{fig:mubrain_architecture}%
    \vspace{-10pt}
\end{figure}

\mubrain{} is an asynchronous and fully-synthesizable digital inference hardware designed in 40nm CMOS~\cite{mubrain}. Figure~\ref{fig:mubrain} shows the internal architecture of a \mubrain{} design with three layers, which are referred to as \texttt{l0, l1}, and \texttt{l2}. There are 336 neurons in the chip, which are of integrate-and-fire (IF) type~\cite{ifneuron}. Figure~\ref{fig:state_diagram} shows the state transitions in an IF neuron in \mubrain{}. 
Every neuron independently (without a global clock) accumulates
weighted incoming synaptic spikes and emits a spike itself
when the neuron's accumulator overflows. Input spikes trigger
the membrane voltage integration, with immediate threshold
evaluation, resulting in distributed granular activations.
Synaptic memory is tightly integrated in the design and distributed closer to neurons, minimizing data (spike) movement.
Static power of the design is \ineq{40.3\mu W} and the dynamic energy per spike is \ineq{26pJ}~\cite{mubrain}.

Synaptic connections between the three layers are fully-programmable, allowing implementation of SDCNN operations such as convolution, pooling, concatenation and addition, as well as irregular network topologies, which are commonly found in many emerging SDCNN models (see Figure~\ref{fig:cnn_structure}). 

\mubrain{} can be scaled up to implement larger network topologies. However, static power and design area increases substantially for larger configurations as reported in Figure~\ref{fig:mubrain}.
To improve energy efficiency, we propose a scalable many-core design where each core is a tiny-scale \mubrain{} (see Section~\ref{sec:mubrain_many_core}).

\subsection{Other Neuromorphic Hardware Designs}
Table~\ref{tab:hw_examples} shows the capacity of recent neuromorphic designs. Here, we review two representative designs -- DYNAPs and Loihi, and show that the proposed \tech{} framework can be easily extended to map SDCNN applications on them.

\vspace{-15pt}
\begin{table}[h!]
	\renewcommand{\arraystretch}{0.8}
	\setlength{\tabcolsep}{1pt}
	\caption{Capacity of recent neuromorphic hardware platforms~\cite{benmezianehardware,lee2020development,schuman2017survey,catthoor2018very,ankita_igsc}.}
	\label{tab:hw_examples}
	\vspace{-5pt}
	\centering
	\begin{threeparttable}
	{\fontsize{6}{10}\selectfont
		\begin{tabular}{c|cccccccc}
			\hline
			& \textbf{ODIN} & $\mathbf{\mu}$\textbf{Brain} & \textbf{DYNAPs} & \textbf{BrainScaleS} & \textbf{SpiNNaker} & \textbf{Neurogrid} & \textbf{Loihi} & \textbf{TrueNorth}\\
			\hline
			\textbf{\# Neurons/core} & 256 & 336 & 256 & 512 & 36K & 65K & 130K & 1M\\
			\textbf{\# Synapses/core} & 64K & 38K & 16K & 128K & 2.8M & 8M & 130M & 256M\\
			\textbf{\# Cores/chip} & 1 & 1 & 1 & 1 & 144 & 128 & 128 & 4096\\
			\hline
			\textbf{\# Chips/board} & 1 & 1 & 4 & 352 & 56 & 16 & 768 & 4096\\
 			\hline
 			\textbf{\# Neurons} & 256 & 336 & 1K & 4M & 2.5B & 1M & 100M & 4B\\
			\textbf{\# Synapses} & 64K & 38K & 65K & 1B & 200B & 16B & 100B & 1T\\
			\hline
	\end{tabular}}
	\end{threeparttable}
\end{table}
\vspace{-5pt}

DYNAPs~\cite{dynapse} is a mixed-signal inference hardware with four neurosynaptic cores. Each core can map up to 256 neurons and 16K synaptic connections. Cores are interconnected using a hierarchical NoC with mesh routing. 
Weights of a fully-trained network (i.e., the inference) are programmed to the synaptic cells of the four cores. Once programmed, DYNAPs can perform inference on streaming data continuously. 
Each DYNAPs core uses a crossbar where neurons are organized into two layers. Synaptic memory is tightly integrated with neuron circuits as in \mubrain{}.
\sm{}~\cite{spinemap} is used to map applications on DYNAPs.%

Loihi~\cite{loihi} is a digital design of a many-core neuromorphic hardware consisting of 128 cores with mesh routing. A Loihi core can map up to 130K neurons and 130M synaptic connections. Unlike \mubrain{} and DYNAPs, synaptic memory in Loihi is off-chip to the neuron circuitry, leading to higher data movements. 
The LAVA framework~\cite{loihi_mapping} is used to map applications on Loihi.

\subsection{System Software for Neuromorphic Hardware}
Apart from SpiNeMap and LAVA, there are also other system software frameworks for spiking neuromorphic accelerators -- 
neutram~\cite{ji2016neutrams}, neuroxplorer~\cite{neuroxplorer}, pacman~\cite{pacman}, and dfsynthesizer~\cite{dfsynthesizer_pp}, among others~\cite{espine,twisha_energy,ncrtm,corelet}. 
All these frameworks use graph partitioning technique to first partition an SDCNN into clusters and then place these clusters onto homogeneous neuromorphic cores connected in a mesh-based topology. 
They cannot be easily extended to hardware with different core capacities and interconnected using segmented bus interconnect.

\section{Many-Core \mubrain{} Design}\label{sec:mubrain_many_core}
\begin{figure}[h!]
	\centering
	\vspace{-10pt}
	\centerline{\includegraphics[width=0.9\columnwidth]{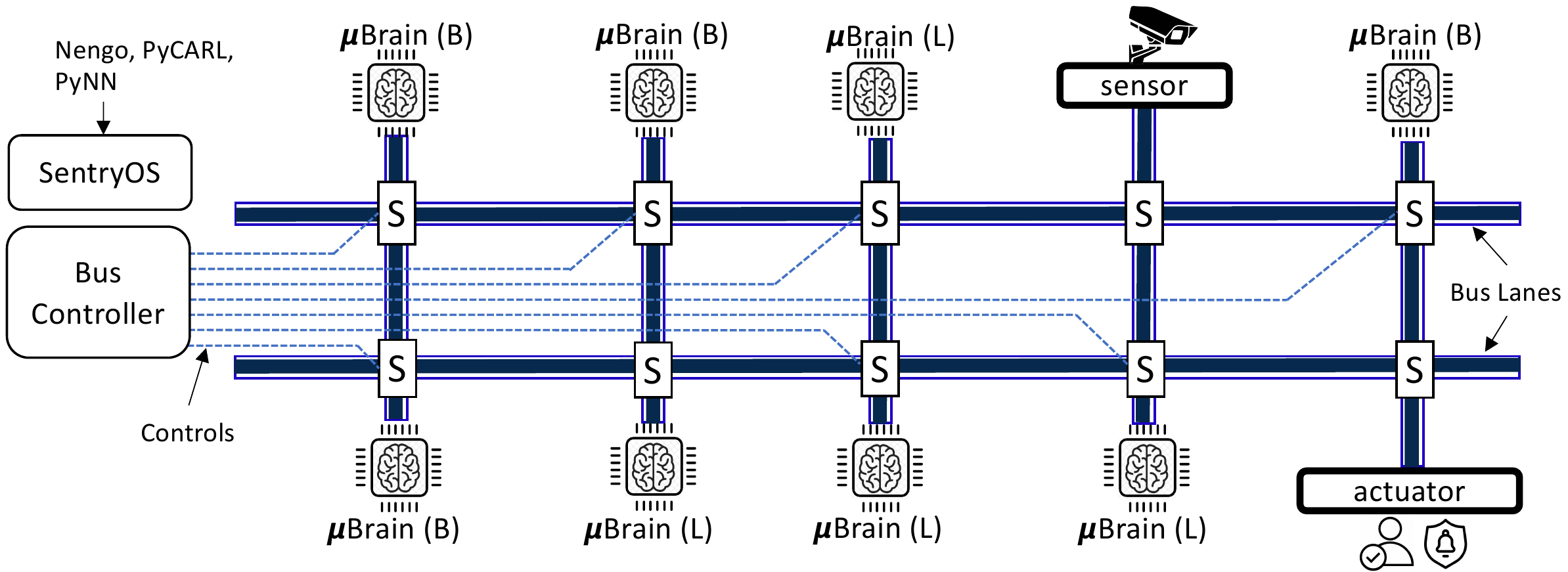}}
	\vspace{-10pt}
	\caption{Many-core big little \mubrain{} with dynamic segmented bus interconnect.}
	\vspace{-10pt}
	\label{fig:many_core_mubrain}
\end{figure}

Figure~\ref{fig:many_core_mubrain} illustrates a high-level architecture of the proposed many-core big little \mubrain{} platform template design, where big (B) and little (L) cores are interconnected using parallel bus lanes that are segmented using segmentation switches (S). 
An SDCNN application (specified using Nengo~\cite{nengo}, PyCARL~\cite{pycarl} or PyNN~\cite{pynn}) is admitted to this platform using the proposed \tech{} framework. A bus controller is used to map inter-core data communication to parallel segments.
\subsection{Big Little \mubrain{} Design}\label{sec:big_little_design_choice}
\begin{figure}[h!]
	\centering
	\vspace{-10pt}
	\centerline{\includegraphics[width=0.99\columnwidth]{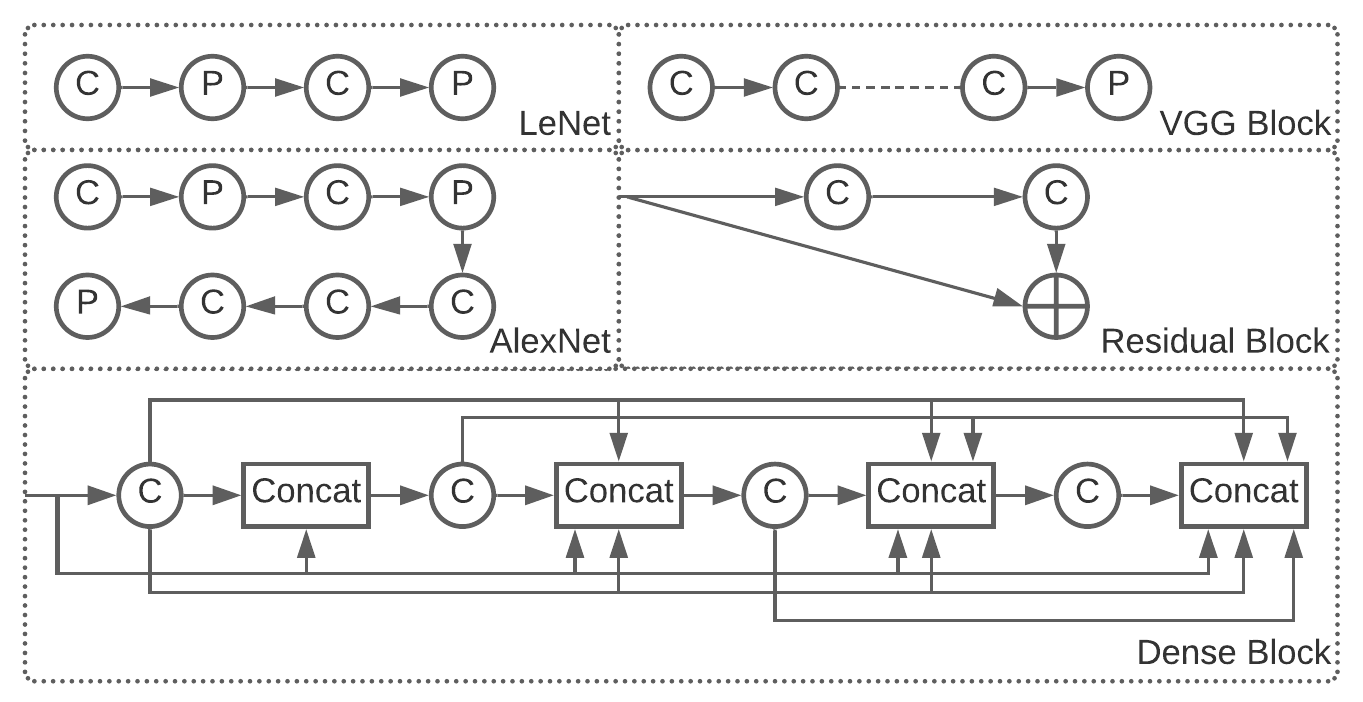}}
	\vspace{-10pt}
	\caption{Structure of five commonly used CNNs with convolution (C), pooling (P), concatenation (concat), and addition (+) operations.}
	\vspace{-5pt}
	\label{fig:cnn_structure}
\end{figure}

Figure~\ref{fig:cnn_structure} illustrates the structure of five commonly-used SDCNNs -- LeNet (1989), AlexNet (2012), VGGNet (2015), ResNet (2015), and DenseNet (2017)~\cite{sengupta2019going}. We observe that LeNet, AlexNet, and VGGNet have a regular topology with chain-like connections. However, emerging CNNs such as ResNet (identity shortcut connections) and DenseNet (one-to-all subsequent layer connections) have an irregular topology~\cite{zheng2020efficient}. Additionally for energy-efficient implementation on embedded devices, SDCNNs are subject to connection pruning, where connections with near-zero synaptic weights are removed~\cite{augustine2019generating}. Such pruning creates an irregular topology, even for LeNet, AlexNet, and VGGNet, making it difficult to map them onto hardware accelerators~\cite{voss2017convolutional,shan2019exact,xu2020autodnnchip,shihao_designflow,shihao_soda}.

To this end, Table~\ref{tab:l1_l2_neighbors} reports the minimum, maximum, and average number of \texttt{L1} and \texttt{L2} neighbors of neurons from these five SDCNNs.\footnote{\texttt{L1} neighbors of a neuron is the set of pre-synaptic neurons that are connected directly to this neuron. \texttt{L2} neighbors of a neuron is the set of pre-synaptic neurons that are connected to \texttt{L1} neighbors of the neuron.} 
We observe that the number of \texttt{L1} and \texttt{L2} neighbors of neurons in an SDCNN varies widely. To use \mubrain{} for these models, the \mubrain{} design needs to be optimally configured to accommodate the maximum number of \texttt{L1} and \texttt{L2} neighbors. This is reported as the conservative \mubrain{} design choice in row 8 of Table~\ref{tab:l1_l2_neighbors}.

\vspace{-15pt}
\begin{table}[h!]
	\renewcommand{\arraystretch}{0.8}
	\setlength{\tabcolsep}{5pt}
	\caption{Statistics of CNN sub-networks~\cite{sengupta2019going}.}
	\label{tab:l1_l2_neighbors}
	\vspace{-5pt}
	\centering
	\begin{threeparttable}
	{\fontsize{6}{10}\selectfont
		\begin{tabular}{c|ccc|ccc}
			\hline
			& \multicolumn{3}{c}{\textbf{\texttt{L1} neighbors}} & \multicolumn{3}{c}{\textbf{L2 neighbors}} \\ \cline{2-7}
			& \textbf{Min} & \textbf{Max} & \textbf{Avg} & \textbf{Min} & \textbf{Max} & \textbf{Avg}\\
			\hline
			LeNet & 18 & 144 & 115.8 & 128 & 400 & 369.3\\
			AlexNet & 9 & 2566 & 41.9 & 11 & 10154 & 544.2\\
			VGGNet & 3 & 288 & 186.3 & 16 & 14772 & 2437.8\\
			ResNet & 12 & 288 & 244.9 & 16 & 1568 & 695.9\\
			DenseNet & 3 & 288 & 162.8 & 16 & 14772 & 1422.3\\
			\hline
			{Conservative \mubrain{} design} & \multicolumn{3}{c|}{\texttt{L1} neurons = 4,096} & \multicolumn{3}{c}{\texttt{L2} neurons = 16,384} \\
			\hline
	\end{tabular}}
	\end{threeparttable}
\end{table}
\vspace{-10pt}

\begin{figure}[h!]
	\centering
	\vspace{-10pt}
	\centerline{\includegraphics[width=0.99\columnwidth]{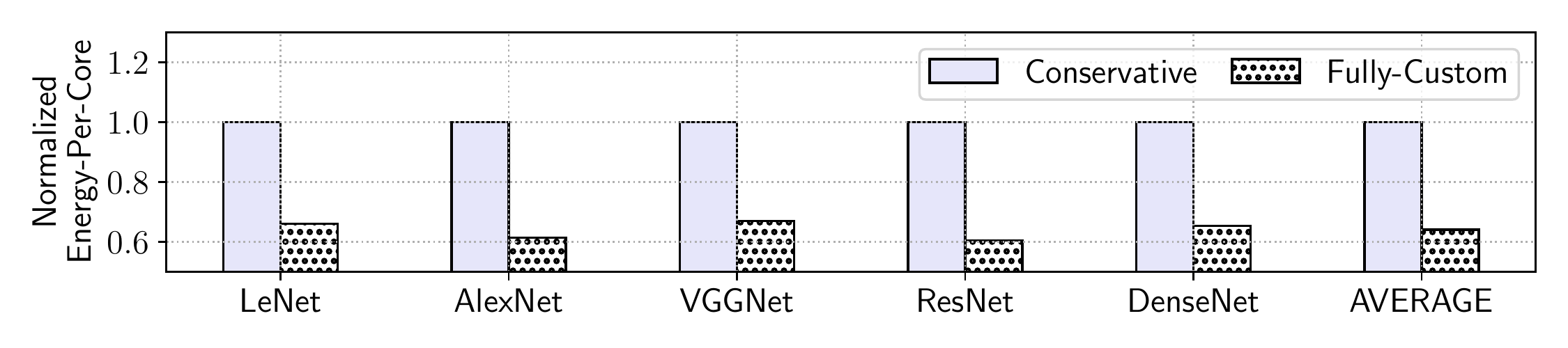}}
	\vspace{-10pt}
	\caption{Average energy-per-core for five SDCNN applications.}
	\vspace{-5pt}
	\label{fig:energy_efficiency}
\end{figure}

Figure~\ref{fig:energy_efficiency} compares the average energy-per-core of the conservative design against a fully-custom design, where \mubrain{} cores are configured based on the number of \texttt{L1} and \texttt{L2} neighbors of neurons for each SDCNN application. Energy numbers 
are normalized to the conservative design. We observe that 
the average energy-per-core of the fully-custom design is on average 36\% lower than the conservative design.
This is because the conservative design is sized based on the maximum number of \texttt{l1} and \texttt{l2} neighbors of a neuron. Such worst-case connectivity occurs only rarely in most applications and therefore, many synaptic connections remain unutilized. This leads to a high static power overhead.
In Section~\ref{sec:mubrain_configs}, we show that using only a few (e.g., 4) \mubrain{} configurations, we can achieve similar energy efficiency as a fully-custom design.

\subsection{Segmented Bus Interconnect}
\begin{figure}[h!]
	\centering
	\vspace{-10pt}
	\centerline{\includegraphics[width=0.99\columnwidth]{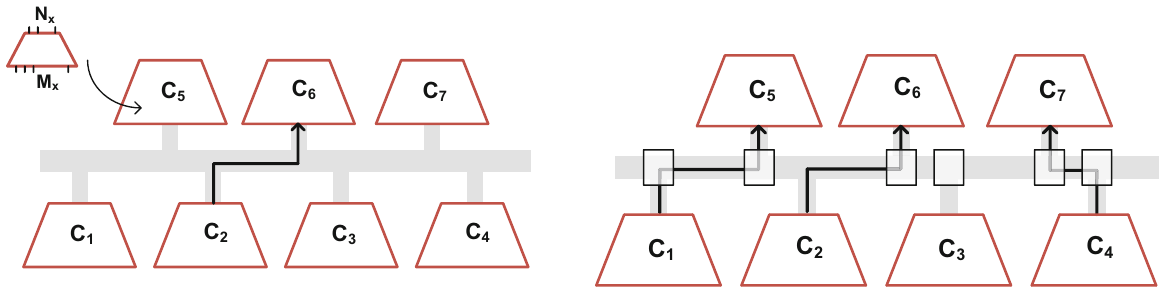}}
	\vspace{-10pt}
	\caption{Segmented bus-based many-core design (right) and its difference with a conventional bus (left).}
	\vspace{-5pt}
	\label{fig:sb_design}
\end{figure}

Figure~\ref{fig:sb_design} illustrates the concept of a segmented bus-based many-core design (right) and its difference with a conventional bus (left) for interconnecting 7 \mubrain{} cores (\ineq{\mathbf{C_1}}-\ineq{\mathbf{C_7}}). Let the core \ineq{\mathbf{C_x}} has \ineq{\mathbf{M_x}} input ports and \ineq{\mathbf{N_x}} output ports. Without loss of generality, we consider a scenario where 
cores \ineq{\mathbf{C_1}}-\ineq{\mathbf{C_4}} can only connect to the inputs of \ineq{\mathbf{C_5}}-\ineq{\mathbf{C_7}}. Fig.~\ref{fig:sb_design} (left) shows a single shared bus connecting \ineq{G = N_1 + N_2 + N_3 + N_4} output ports with \ineq{H = M_5 + M_6 + M_7} input ports. While using shared
bus, only one connection between any pair of input-output cluster is possible at a
given time (shown with the arrow), resulting in underutilization of the
bus. A simple segmented bus allows to overcome this problem by breaking the bus
into multiple segments. As seen from Fig.~\ref{fig:sb_design} (right), a single segmented bus can accommodate many simultaneous connections.

However, a single bus lane may not always be enough to support concurrent connection requests. For instance, any connection request from \ineq{\mathbf{C_3}} to \ineq{\mathbf{C_5}} will be blocked while \ineq{\mathbf{C_2}} is communicating to \ineq{\mathbf{C_6}}.
This can be solved using parallel bus lanes as illustrated in Figure~\ref{fig:many_core_mubrain}.
To achieve full connectivity at all time, \ineq{G*H} separate busses are needed. However, such a full connection is rarely exercised in a
typical SDCNN application. 
A more realistic scenario is the one where at different time
intervals a different connection profile is active. 
We analyze inter-core communications based on training data to identify the minimum number of bus lanes needed in the parallel segmented bus interconnect at any given time.
This minimum value is not dependent on the number of cores.
The key idea here is to exploit the dynamism present in different SDCNN applications to virtualize the segmented bus interconnect between different cores. Hence,
the bandwidth allocation needed at design-time can be reduced from the physical
maximum to what is maximally happening concurrently. Consequently,
the active wire length is reduced, which lowers energy. 
A bus controller is designed to control segmentation switches and map inter-core communications on parallel segments for concurrent execution. This reduces spike latency.

\section{Detailed Design of \tech{}}\label{sec:sentry_ss}
We represent an SDCNN application as a graph. Formally,
\begin{Definition}{SDCNN Graph}
An SDCNN application \ineq{\mathbf{G_{SDCNN} = (\mathbf{N},\mathbf{E})}} is a directed graph consisting of a finite set \ineq{{\mathbf{N}}} of nodes, representing neurons and a finite set \ineq{{\mathbf{E}}} of edges, representing synaptic connections.
\end{Definition}

\begin{figure}[h!]
	\centering
	\vspace{-10pt}
	\centerline{\includegraphics[width=0.99\columnwidth]{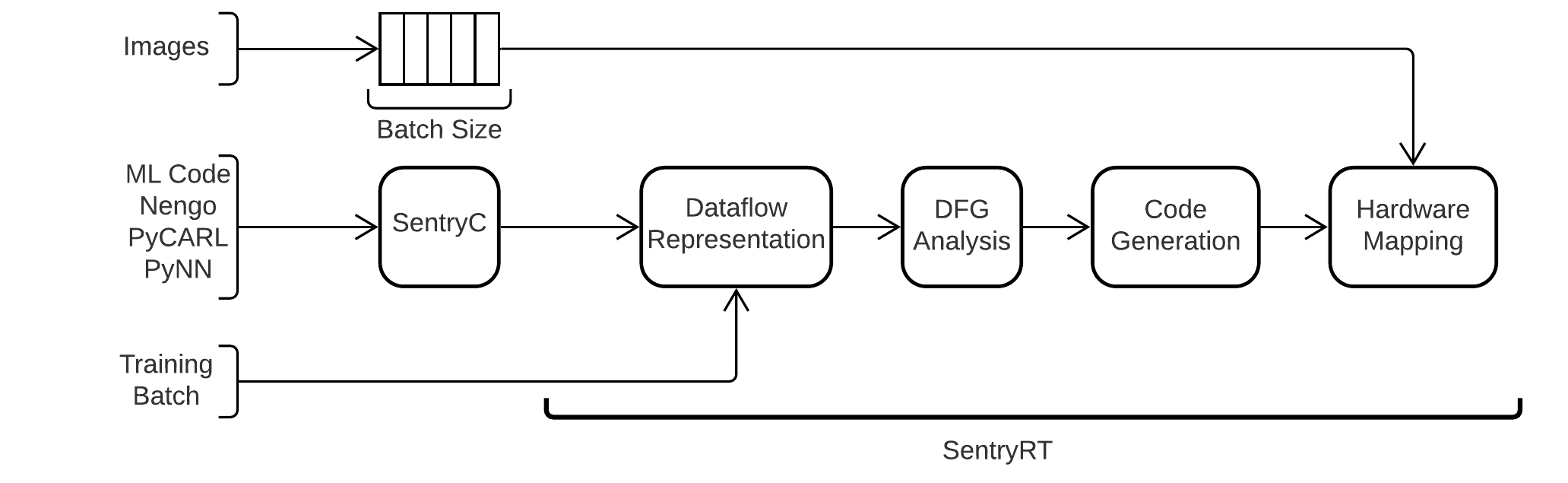}}
	\vspace{-10pt}
	\caption{Detailed architecture of \techrt{}.}
	\vspace{-5pt}
	\label{fig:rtm}
\end{figure}

\tech{} consists of a compiler (\techcompile{}) and a run-time manager (\techrt{}) as illustrated in Figure~\ref{fig:rtm}. 

\subsection{Compiler (\techcompile{}) Design}
\techcompile{} partitions the graph \ineq{G_{SDCNN}}
into sub-networks \ineq{\{S_0,S_1,\cdots\}}, where \ineq{S_i\bigcap S_j = \emptyset}. A sub-network \ineq{S_i = \{\texttt{l2}_i~|~\texttt{l1}_i~|\texttt{l0}_i\}} is a set of neurons that are organized into three layers -- \texttt{l2}, \texttt{l1}, and \texttt{l0}. 
\ineq{Area(S_i)} is the area of the \mubrain{} core needed to implement \ineq{S_i} and \ineq{Power(S_i)} is its static power. Algorithm~\ref{alg:compiler} shows the pseudo-code of \techcompile{}.

\begin{figure}[h!]
	\centering
	\vspace{-10pt}
	\centerline{\includegraphics[width=0.99\columnwidth]{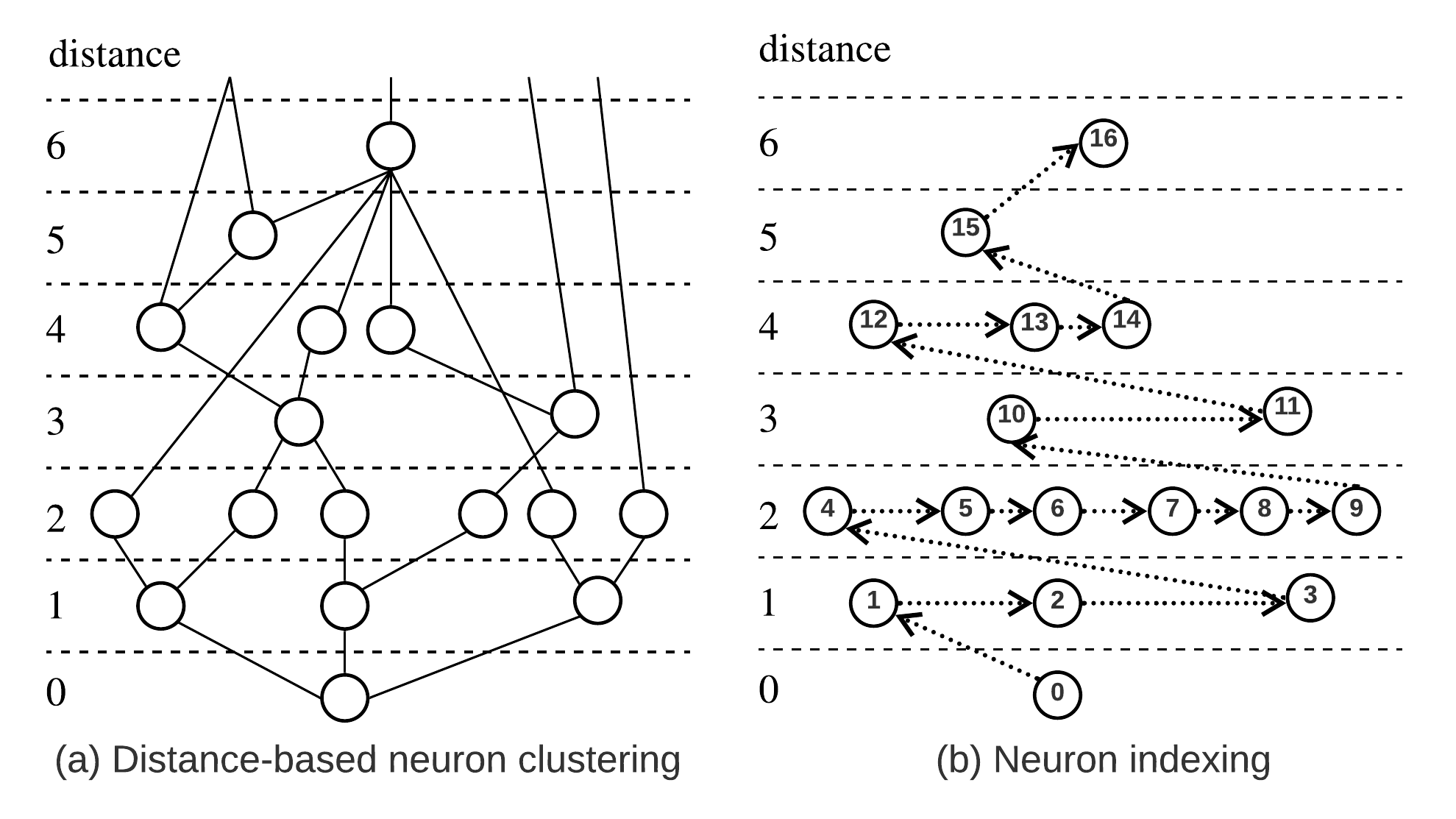}}
	\vspace{-10pt}
	\caption{Distance-based partitioning an SDCNN application into subnetworks.}
	\vspace{-5pt}
	\label{fig:indexing}
\end{figure}

The algorithm operates in four steps.
First, for each output neuron, we assign a distance value to all other neurons, where the distance is computed as
the longest path from a neuron to this output neuron. Neurons that are not connected to this output neuron are assigned a large number (to prevent them from grouping in the same sub-network). Neurons with the same distance are clustered together as shown in Figure~\ref{fig:indexing}a. Second, we index each neuron by the distance sequence as shown in Figure~\ref{fig:indexing}b and form a search path for partitioning the graph into sub-networks: each sub-network must contain neurons with contiguous indexes. In this manner, any arbitrary network topology is transformed into a linear one.

Third, we cluster neurons with distance of up to 2 into a sub-network, i.e., \ineq{S_0 = \{N_j~|~dist(N_j) \leq 2\}}. This is to ensure that each sub-network can fit into the three-layered architecture of \mubrain{}. We compute the remaining graph as \ineq{G' = G - \{S_0\}}.
Next, we recursively partition the remaining graph \ineq{G'} for each node with distance 2 as the new output neuron and compute distance of all other neurons in \ineq{G'} by traversing backward.

\begin{figure}[h!]%
    \centering
    \vspace{-20pt}
    \subfloat[Four sub-networks of the SDCNN example of Figure~\ref{fig:indexing}.\label{fig:subnets}]{{\includegraphics[width=3.7cm]{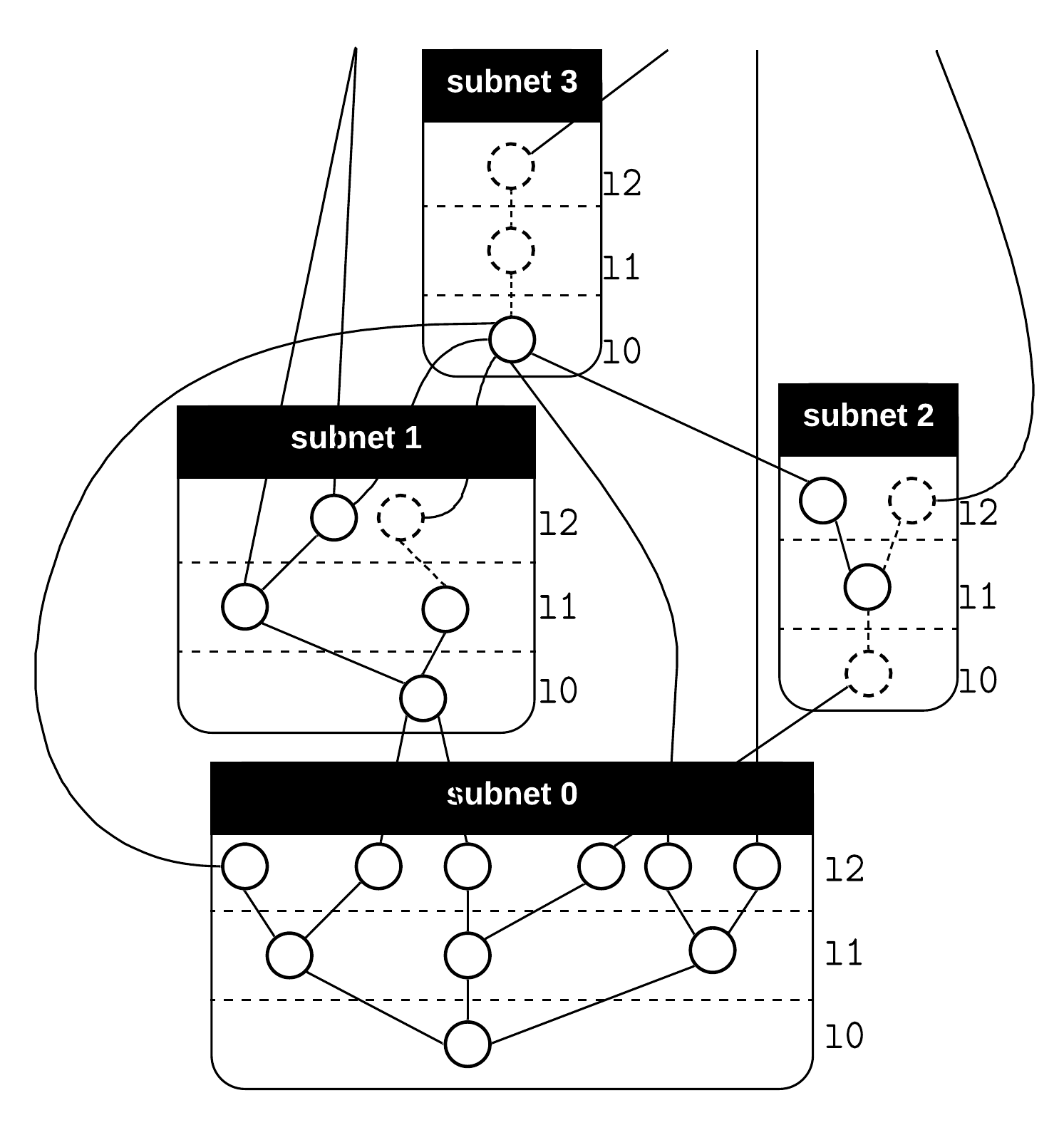} }}%
    \hfill
    \subfloat[Scheduling sub-networks to \mubrain{} pipelines.\label{fig:lenet_mubrain_mapping}]{{\includegraphics[width=5.0cm]{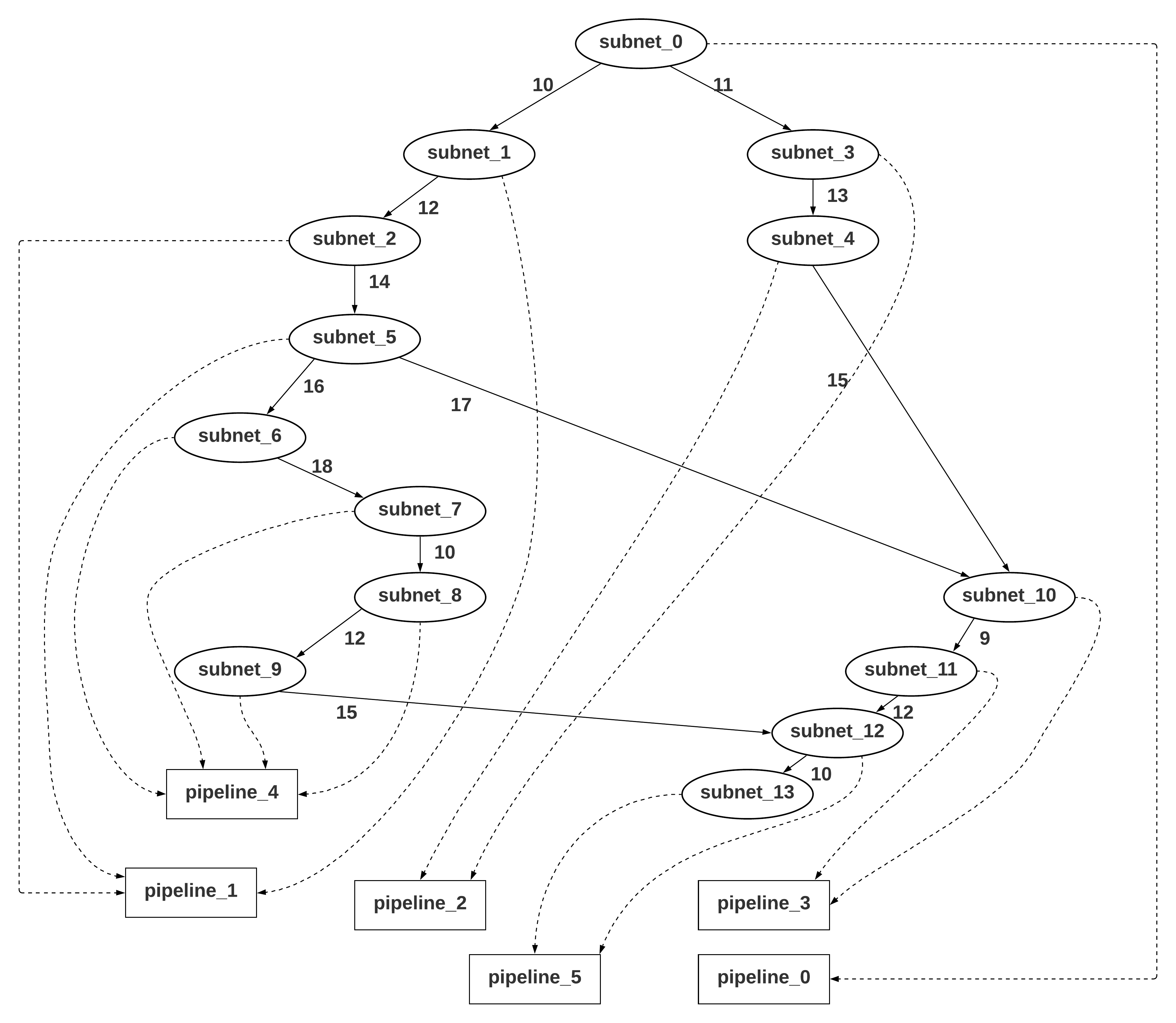} }}%
    \vspace{-5pt}
    \caption{(a) Generating sub-networks by exploiting area-power benefits and (b) scheduling sub-networks of an SDCNN application to \mubrain{} pipelines by exploiting data-level parallelism.}%
    \label{fig:mubrain_subnet}%
    \vspace{-10pt}
\end{figure}

Figure~\ref{fig:subnets} shows the four sub-networks of the SDCNN application of Figure~\ref{fig:indexing}. In generating a sub-network, it may be necessary to create neurons with unit synaptic strength in order to fit onto the three-layer architecture of \mubrain{}. This is shown with the dotted neurons and connections inside sub-networks 1, 2, and 3. 

Finally, we merge sub-networks (lines 7-12 of Algorithm~\ref{alg:compiler}) by considering area and power benefits of big and little \mubrain{} cores. We formulate this problem as follows. Let \ineq{S_i} and \ineq{S_j} be two sub-networks of an SDCNN application. Formally, the merged sub-network is represented as \ineq{S_{i,j} = \{S_{i,j}.\texttt{l2}~|~S_{i,j}.\texttt{l1}~|~S_{i,j}.\texttt{l0}\}}, where \ineq{S_{i,j}.\texttt{l2} = \{S_{i}.\texttt{l2} \cup S_{j}.\texttt{l2}\}}, and so on. We merge \ineq{S_i} and \ineq{S_j} iff
\begin{itemize}
    \item \ineq{{Area(S_{i,j})} < {Area(S_{i})} + {Area(S_{j})}}
    \item \ineq{{Power(S_{i,j})} < {Power(S_{i})} + {Power(S_{j})}}
\end{itemize}

\subsection{Extension of \techcompile{} to Other Spiking Architectures}\label{sec:extension}
\subsubsection{\underline{Extension to DYNAPs}} DYNAPs is a crossbar-based two-layer architecture with fixed number of \texttt{l0} and \texttt{l1} neurons. To use \techcompile{} for DYNAPs, we make the following changes. First, we 
construct sub-networks with neurons that are at a distance 0 (layer 0) and 1 (layer 1) only, with a constraint on the size of a crossbar. This is accomplished by changing line 6 of Algorithm~\ref{alg:compiler} to \ineq{S_i = \texttt{create\_subnet}\left(\underset{d(n_j)\leq 1}{\bigcup} n_j \text{ such that } |S_i| \leq C_n\right)}, where \ineq{C_n} is the total number of neurons that a crossbar can map. Second, we place all neurons that are at a distance of 2 or higher as candidates for sub-network generation in the next iteration. To do so, we change line 14 to \ineq{newNS.append(\underset{d(n_j) = 2}{\bigcup} n_j)}. Finally, we assign all unassigned neurons from the current iteration and the new candidates to the list \ineq{NS} for the next iteration of the algorithm, i.e., change line 16 to \ineq{NS = \{NS\setminus G'\} \bigcup newNS}.

\subsubsection{\underline{Extension to Loihi}} 
To use \techrt{} for Loihi, where neuron circuitry is decoupled from synaptic memory, we propose the same changes as DYNAPs with two constraints -- one on the number of neurons and the other on the size of synaptic memory. Additionally, while placing a sub-network to a core, we decouple the neurons from the synapses and place them separately onto the target core architecture.

\vspace{-10pt}
\begin{algorithm}[h]
	\scriptsize{
 		\KwIn{\ineq{G_{SDCNN}= (\textbf{N},\textbf{E})}}
 		\KwOut{\ineq{G_{DFG}= (\textbf{S},\textbf{C})}}
 		$G' = G_{SDCNN}$\tcc*[r]{Initialize $G'$}
 		$NS = \{\underset{\texttt{OutDegree}(n_i) = 0}{\bigcup} n_i\}$\tcc*[r]{Place all output neurons.}
 		\While(\tcc*[f]{Run until all neurons of the graph are placed into sub-networks}){$G'\neq \emptyset$}{
 		    \For(\tcc*[f]{For each neurons in $NS$}){$n_i\in NS$}{
 		        $d(n_j) = \texttt{longest\_path}(n_j,n_i)~~\forall~n_j\in G'$\tcc*[r]{Compute distance of every neuron in $G'$ to $n_i$}
 		        $S_i = \texttt{create\_subnet}\left(\underset{d(n_j)\leq 2}{\bigcup} n_j\right)$\tcc*[r]{Create subnet with all neurons that have a distance of 2 or less from the output neuron}
 		        
 		         Find $S_j~|~j = \texttt{argmin}\{Cost(S_k,S_i)~\forall~S_k\in G_{DFG}\}$\tcc*[r]{Find one subnet amongst the created ones to which merging this new subnet results in the minimum cost. Cost can be assessed in terms of area, power, or a combination of the two, which we have used in this paper.}
 		         \uIf(\tcc*[f]{If the area and power costs of merging the subnets is less than the individual costs.}){$Area(S_{i,j}) < Area(S_i) + Area(S_j)$ \textbf{ and } $Power(S_{i,j}) < Power(S_i) + Power(S_j)$}{
 		            $S_j = S_j \bigcup S_i$\tcc*[r]{Merge the subnets.}
 		         }
 		         \Else{ $G_{DFG}.\texttt{append}(S_i)$\tcc*[r]{Insert this subnet to the output DFG}
 		        }
 		        $G' = G' - S_i$\tcc*[r]{Remove $S_i$ from $G'$}
 		        $newNS.append(\underset{d(n_j) = 3}{\bigcup} n_j)$\tcc*[r]{Use the set $newNS$ to hold all neurons that are at a distance of 3. Neurons in this set will be placed in the next iteration.}
 		    }
 		    $NS = newNS$\tcc*[r]{Update $NS$ to $newNS$.}
 		}
 	}
	\caption{Compiler (\techrt{}) of \tech{}.}
	\label{alg:compiler}
\end{algorithm}
\vspace{-15pt}

\subsection{Run-time Manager (\techrt{}) Design}
Figure~\ref{fig:rtm} shows the proposed \techrt{}. It queues input images to a given batch size and process them concurrently.
Although queuing increases latency, the throughput is higher due to batch processing.
\techrt{} uses a dataflow analysis technique to schedule sub-networks onto \mubrain{} cores to improve throughput. To this end, \techrt{} uses a training batch to profile an SDCNN application and represent its sub-networks and their interconnections as a dataflow graph (DFG). Formally,
\begin{Definition}{SDCNN DFG Graph}
An SDCNN dataflow graph \ineq{\mathbf{G_{DFG} = (\mathbf{S},\mathbf{C})}} is a directed graph consisting of a finite set \ineq{{\mathbf{S}}} of sub-networks of the SDCNN application and a finite set \ineq{{\mathbf{C}}} of communication channels between the sub-networks.
\end{Definition}
Each sub-network \ineq{S_i\in \mathbf{S}} is associated with an execution time \ineq{t_i}, which represents its computation time on a \mubrain{} core.

\begin{figure}[h!]
	\centering
	\vspace{-10pt}
	\centerline{\includegraphics[width=0.99\columnwidth]{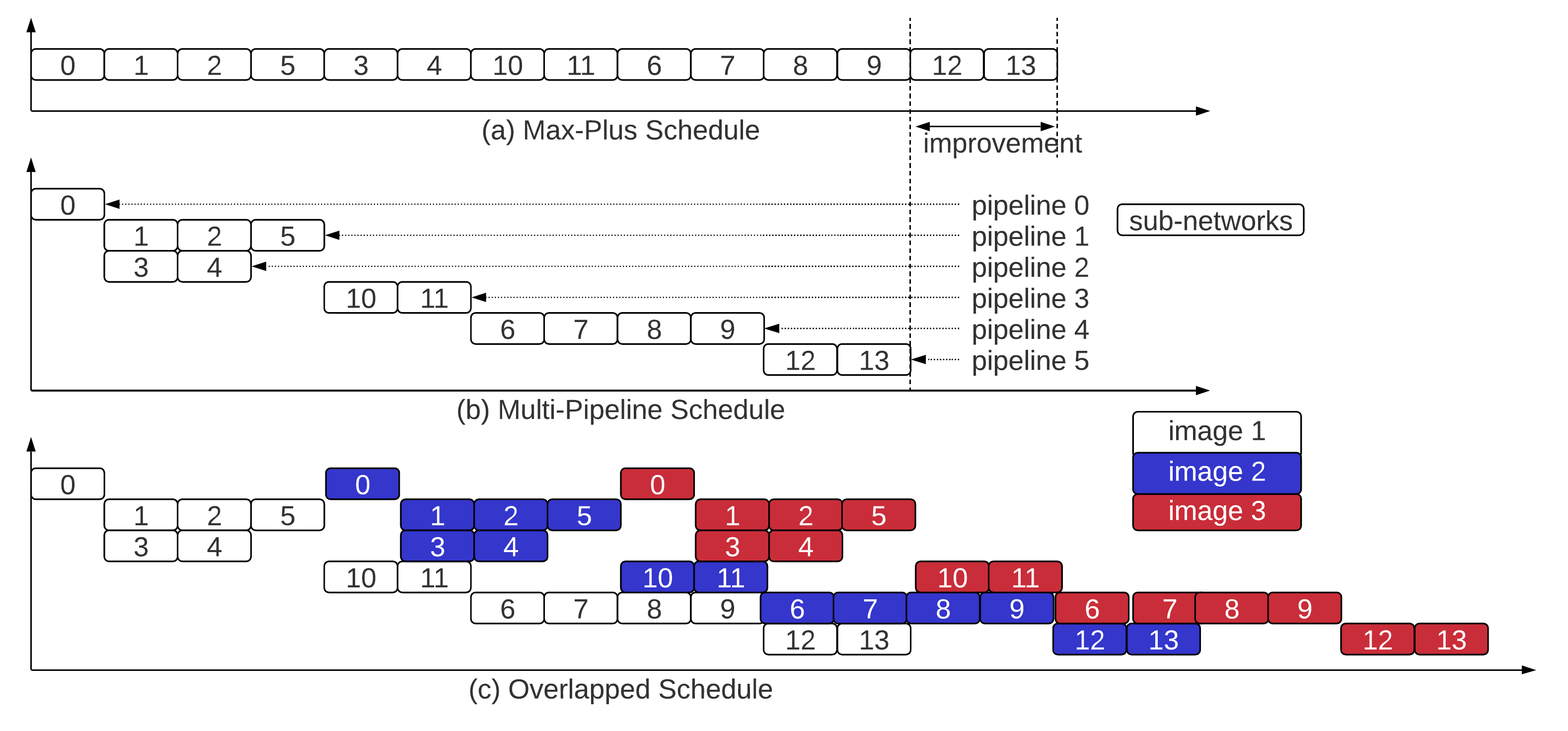}}
	\vspace{-10pt}
	\caption{Schedule of sub-networks generated by \techrt{}.}
	\vspace{-5pt}
	\label{fig:schedule}
\end{figure}

\techrt{} uses an analytical approach to timing analysis of \ineq{G_{DFG}}~\cite{thiele2000real}. 
It consists of a novel way of constructing a Max-Plus algebraic description of the evolution of node execution times in a self-timed execution manner.
The Max-Plus semiring \ineq{\mathbb{R}_{\text{max}}} is the set \ineq{\mathbb{R}\cup\{-\infty\}} defined with two basic operations \ineq{\oplus \text{ and } \otimes}, which are related to linear algebra as \ineq{a \oplus b = \max(a,b) \text{  and  } a \otimes b = a + b}.
The identity element \ineq{\mymathbb{0}} for the addition \ineq{\oplus} is \ineq{-\infty} in linear algebra, i.e., \ineq{a \oplus \mymathbb{0} = a}. The identity element \ineq{\mymathbb{1}} for the multiplication \ineq{\otimes} is 0 in linear algebra, i.e., \ineq{a \otimes \mymathbb{1} = a}.
The end execution time of each node of \ineq{G_{DFG}} can be expressed as \ineq{\mathbf{t_k} = \oplus\mathbf{{T}\otimes t_{k-1}}},
where $\mathbf{{T}}$ captures execution times of sub-networks $\tau_{n}$ and \ineq{\mathbf{t_k} = \{t_0(k),t_1(k),\cdots\}} is the end execution of nodes in the \ineq{k^{th}} iteration.
Figure~\ref{fig:schedule}a shows the schedule obtained by solving the Max-Plus formulation of end execution time of sub-networks of an SDCNN micro-benchmark shown in Figure~\ref{fig:lenet_mubrain_mapping}.

To exploit parallelism, \techrt{} discards the precise execution time 
of sub-networks, retaining only their sequence. Sub-networks are executed in a self-timed fashion when data is available for its neurons to compute.
Figure~\ref{fig:lenet_mubrain_mapping} shows the allocation of sub-networks to 6 \mubrain{} pipelines, where each pipeline may consist of a few \mubrain{} cores. 
The corresponding multi-pipeline schedule is illustrated in Figure~\ref{fig:schedule}b. Finally,
\techrt{} overlaps execution of pipelines for different images of a batch to improve throughput.
Figure~\ref{fig:schedule}c shows this overlapped execution for three different input images.

\section{Evaluation}\label{sec:evaluation}
\mubrain{} design is synthesized at 40 nm technology node using TSMC library CLN40LP. Area and power numbers for different \mubrain{} configurations are estimated using the compact model provided in~\cite{mubrain}. We use an in-house cycle-accurate neuromorphic platform simulator to simulate many-core \mubrain{} cores interconnected using a segmented bus interconnect. We use PyCARL~\cite{pycarl} to simulate five commonly-used SDCNN applications with 2-bit quantized synaptic weights. These applications are described in Table~\ref{tab:apps}.

\vspace{-10pt}
\begin{table}[h!]
	\renewcommand{\arraystretch}{0.8}
	\setlength{\tabcolsep}{2pt}
	\caption{SDCNN applications used to evaluate the proposed design.}
	\label{tab:apps}
	\vspace{-5pt}
	\centering
	\begin{threeparttable}
	{\fontsize{6}{10}\selectfont
		\begin{tabular}{cc|ccc|c}
			\hline
			 \textbf{SDCNN} &
			\textbf{Dataset} &
			\textbf{Neurons} & \textbf{Synapses} & \textbf{Avg. Spikes/Image} & \textbf{Accuracy}\\
			\hline
			LeNet & CIFAR-10 & 80,271 & 275,110 & 724,565 & 86.3\%\\
			AlexNet & CIFAR-10 & 127,894 & 3,873,222 & 7,055,109 & 66.4\%\\
			VGGNet & CIFAR-10 & 448,484 & 22,215,209 & 12,826,673 & 81.4 \%\\
			ResNet & CIFAR-10 & 266,799 & 5,391,616 & 7,339,322 & 57.4\%\\
			DenseNet & CIFAR-10 & 365,200 & 11,198,470 & 1,250,976 & 46.3\%\\
			\hline
	\end{tabular}}
	\end{threeparttable}
\end{table}
\vspace{-10pt}

\subsection{Energy Efficiency}
\begin{figure}[h!]
	\centering
	\vspace{-10pt}
	\centerline{\includegraphics[width=0.99\columnwidth]{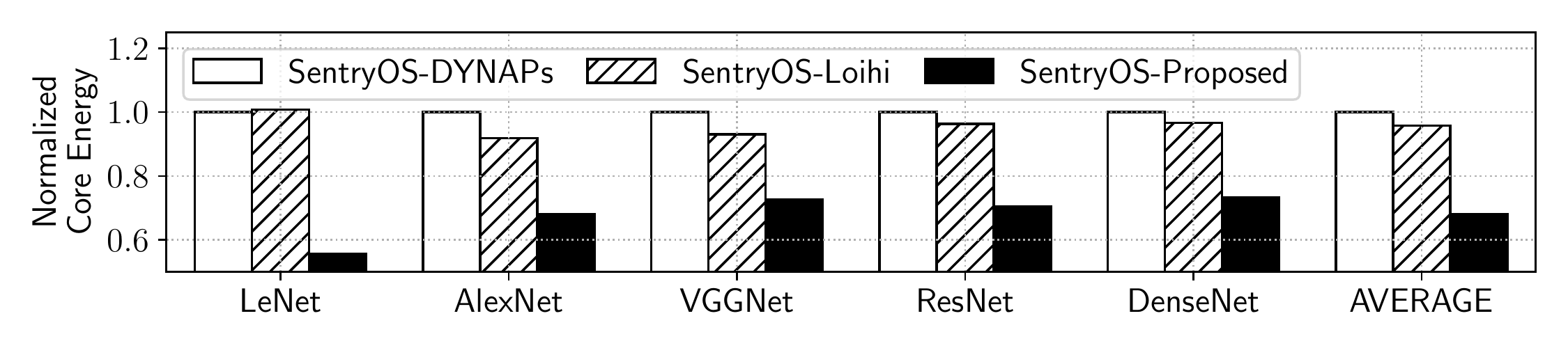}}
	\vspace{-10pt}
	\caption{Core energy normalized to DYNAPs.}
	\vspace{-5pt}
	\label{fig:core_energy}
\end{figure}

Figure~\ref{fig:core_energy} plots the core energy of DYNAPs, Loihi, and the proposed many-core \mubrain{} design for all five SDCNN applications. We scale DYNAPs and Loihi to 40nm node (the same technology node as \mubrain{}). We use the proposed \tech{} for all these three spiking neuromorphic accelerators. Results are normalized to DYNAPs. We make two key observations.

First, energy of Loihi is on average 5\% lower than DYNAPs. Although Loihi requires higher energy to access off-chip synaptic memory, the improvement is due to larger capacity of a Loihi core compared to DYNAPs. For smaller applications like LeNet, energy numbers are comparable. Second, energy using the proposed design is on average 32\% lower than DYNAPs and 29\% lower than Loihi. This improvement is because 1) a \mubrain{}  core consumes lower power than a DYNAPs and Loihi core~\cite{mubrain}, 2) contrary to DYNAPs and Loihi, the proposed design uses different capacity (big little) \mubrain{} cores, which significantly reduces unused synaptic connections and improves energy efficiency, 3) the proposed design uses parallel segmented bus interconnect, which is more energy efficient than a traditional mesh-based NoC, which is used in DYNAPs and Loihi, and 4) contrary to Loihi, where synaptic memory is separated from neurons, \mubrain{} requires very low data movement because of the integration of synaptic memory with neurons.

\subsection{Throughput}

\begin{figure}[h!]
	\centering
	\vspace{-10pt}
	\centerline{\includegraphics[width=0.99\columnwidth]{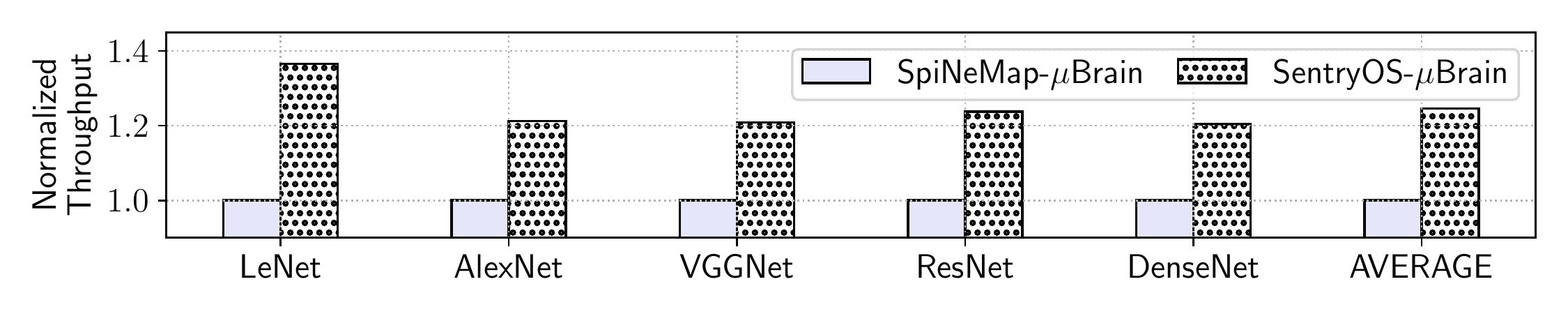}}
	\vspace{-10pt}
	\caption{Throughput normalized to SpiNeMap.}
	\vspace{-5pt}
	\label{fig:throughput}
\end{figure}

Figure~\ref{fig:throughput} plots throughput of the evaluated SDCNN applications on the proposed design. We compare \tech{} with a previously-proposed framework \sm{}. Results are normalized to \sm{}. We make two observations.

First, throughput using \tech{} is on average 25\% higher than \sm{}. This improvement is because, \tech{} first compiles an SDCNN into sub-networks by exploiting the internal architecture of big and little \mubrain{} cores and then uses a dataflow analysis technique to schedule sub-networks to cores improving opportunities for pipelining and exploiting data-level parallelism. Second, even for irregular topologies such as ResNet and DenseNet, \tech{} results in an average 22\% higher throughput than \sm{}.

\subsection{\mubrain{} Design Choice: Segmented Bus Interconnect}\label{sec:energy}
\begin{figure}[h!]%
    \centering
    \vspace{-10pt}
    \subfloat[Normalized interconnect energy.\label{fig:energy}]{{\includegraphics[width=8.0cm]{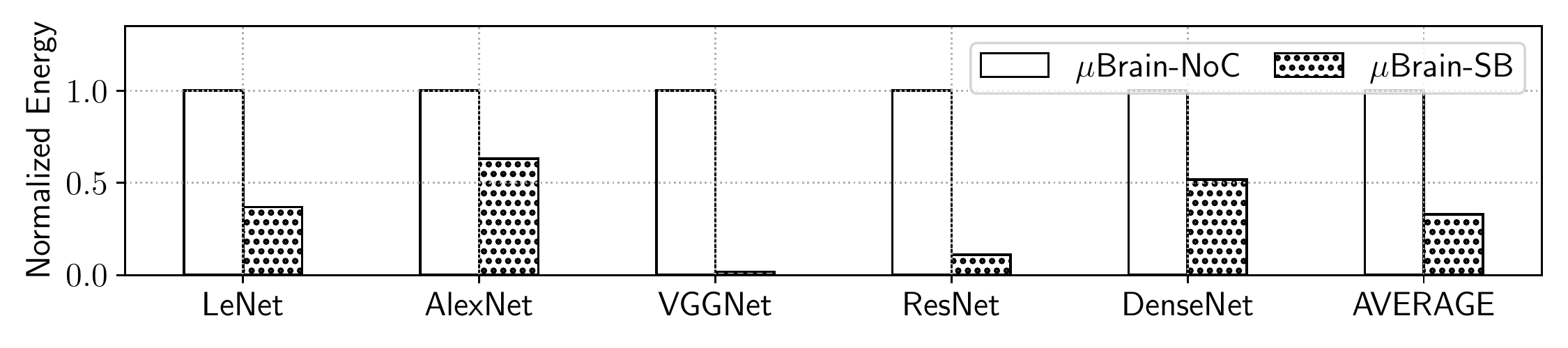} }}%
    \hfill
    \subfloat[Normalized latency.\label{fig:latency}]{{\includegraphics[width=8.0cm]{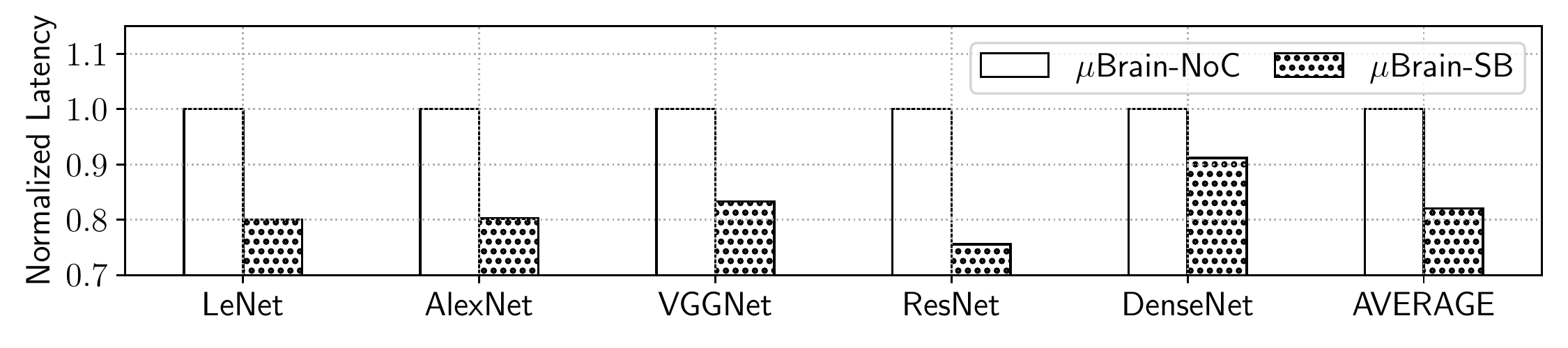} }}%
    \vspace{-5pt}
    \caption{Comparing \mubrain{} design with segmented bus (SB) interconnect to mesh-based network-on-chip (NoC) interconnect.}%
    \label{fig:mubrain_energy_latency}%
    \vspace{-10pt}
\end{figure}

Figure~\ref{fig:mubrain_energy_latency} plots interconnect energy and latency of the proposed many-core design with segmented bus (SB) compared to a mesh-based network-on-chip (NoC) interconnect. We make the following two key observations.

First, energy using segmented bus interconnect is on average 67\% lower than NoC. This improvement is because 1) we analyze inter-core communication based on training data to identify the minimum number of parallel bus lanes needed in the segmented bus interconnect, which reduces the active wire length compared to NoC, 
and 
2) core-to-core communications do not need to wake and utilize the entire bus; rather, only segments connecting the communicating cores need to be powered up.
Second, latency using segmented bus is on average 18\% lower than that of NoC. This is because in the proposed design, segmentation switches are programmed once at design-time before admitting an application. This is done by analyzing the communication profile. Since there is no run-time routing decisions involved, latency is lower.

\subsection{\mubrain{} Design Choice: Heterogeneous Configurations}\label{sec:mubrain_configs}
\begin{figure}[h!]
	\centering
	\vspace{-10pt}
	\centerline{\includegraphics[width=0.99\columnwidth]{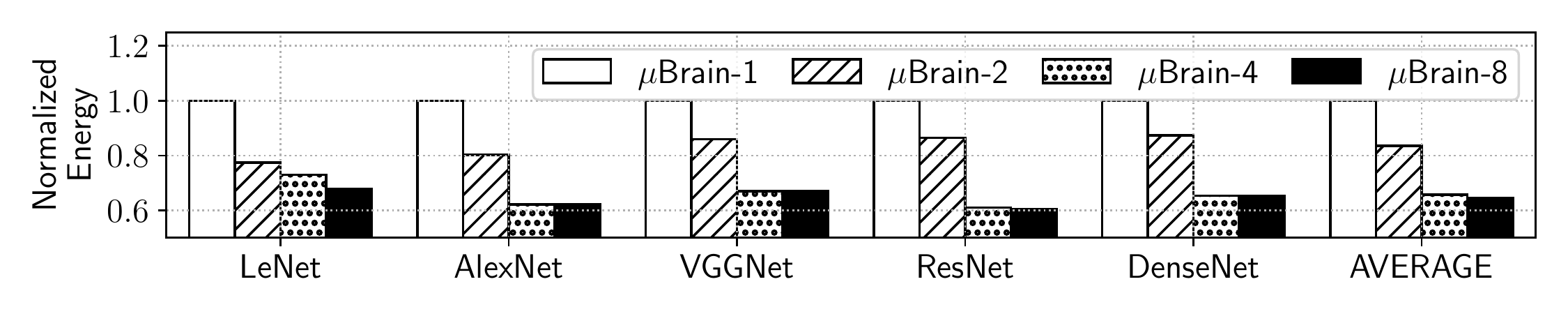}}
	\vspace{-10pt}
	\caption{Energy for different \mubrain{} configurations.}
	\vspace{-5pt}
	\label{fig:heterogeneous}
\end{figure}

Figure~\ref{fig:heterogeneous} plots energy of the proposed \mubrain{}-based many-core design with 1, 2, 4, and 8 different \mubrain{} configurations. The proposed design with a single configuration is the conservative design of Table~\ref{tab:l1_l2_neighbors}. All results are normalized to this conservative design. We make two observations.

First, with 2, 4, and 8 configurations, energy is on average 16\%, 35\%, and 36\% lower than the conservative design, respectively. In comparison, energy of a fully-custom SDCNN-specific design, where cores can be of difference sizes, is 36\% lower than the conservative design (see Fig.~\ref{fig:energy_efficiency}). Energy reduces with more configurations due to the reduction of unused synaptic connections in each core. 
Second, increasing from 4 to 8 configurations, the reduction of energy is less than 1\%. In our proposed design template, we have used only four configurations  -- 1) little, type 1 \ineq{(256\times64\times16)}, 2) little, type 2 \ineq{(1024\times256\times16)}, 3) big, type 1 \ineq{(4096\times1024\times16)}, and 4) big. type 2 \ineq{(16384\times4096\times16)}, instead of adopting a fully-custom SDCNN-specific design. This is to make the design generic and applicable to many different SDCNN inference applications.

\section{Conclusion}\label{sec:conclusion}
We introduce a many-core neuromorphic platform template consisting of asynchronous (clock-less) big little digital \mubrain{} cores interconnected using a segmented bus interconnect.
We propose a system software framework \tech{}, consisting of a compiler and a run-time manager, to compile spiking deep convolutional neural network (SDCNN) to the proposed design. 
Using five commonly-used SDCNN applications, we show a significant improvement in energy, latency, and throughput. 



\bibliographystyle{IEEEtran}
\bibliography{commands,disco,external}

\end{document}